\documentclass[10pt,twocolumn,letterpaper]{article}

\usepackage{cvpr}              

\usepackage[dvipsnames]{xcolor}

\usepackage{adjustbox}
\definecolor{cvprblue}{rgb}{0.21,0.49,0.74}
\usepackage[pagebackref,breaklinks,colorlinks,citecolor=cvprblue]{hyperref}
\usepackage[accsupp]{axessibility} 
\def\paperID{2476} 
\def\confName{CVPR}
\def\confYear{2024}

\title{Scene Adaptive Sparse Transformer for Event-based Object Detection}

\author{
    Yansong Peng\textsuperscript{1,*}\and
    Hebei Li\textsuperscript{1,*} \and
    Yueyi Zhang\textsuperscript{1,\dag} \and
    Xiaoyan Sun\textsuperscript{1,2} \and
    Feng Wu\textsuperscript{1,2} \and
\textsuperscript{1}University of Science and Technology of China \and
\textsuperscript{2}Institute of Artificial Intelligence, Hefei Comprehensive National Science Center\\
{\tt\small \{pengyansong, lihebei\}@mail.ustc.edu.cn, 
\{zhyuey, sunxiaoyan, fengwu\}@ustc.edu.cn
}
}

\begin{document}
\maketitle

\let\thefootnote\relax\footnotetext{\textsuperscript{*}Joint first author \textsuperscript{\dag}Corresponding author}
\begin{abstract}
While recent Transformer-based approaches have shown impressive performances on event-based object detection tasks, their high computational costs still diminish the low power consumption advantage of event cameras. Image-based works attempt to reduce these costs by introducing sparse Transformers. However, they display inadequate sparsity and adaptability when applied to event-based object detection, since these approaches cannot balance the fine granularity of token-level sparsification and the efficiency of window-based Transformers, leading to reduced performance and efficiency. Furthermore, they lack scene-specific sparsity optimization, resulting in information loss and a lower recall rate. To overcome these limitations, we propose the Scene Adaptive Sparse Transformer (SAST). SAST enables window-token co-sparsification, significantly enhancing fault tolerance and reducing computational overhead. Leveraging the innovative scoring and selection modules, along with the Masked Sparse Window Self-Attention, SAST showcases remarkable scene-aware adaptability: It focuses only on important objects and dynamically optimizes sparsity level according to scene complexity, maintaining a remarkable balance between performance and computational cost. The evaluation results show that SAST outperforms all other dense and sparse networks in both performance and efficiency on two large-scale event-based object detection datasets (1Mpx and Gen1). 

\noindent{Code:  \url{https://github.com/Peterande/SAST}.}

\end{abstract}

\begin{figure}[t]
\centering
\includegraphics[width=\columnwidth]{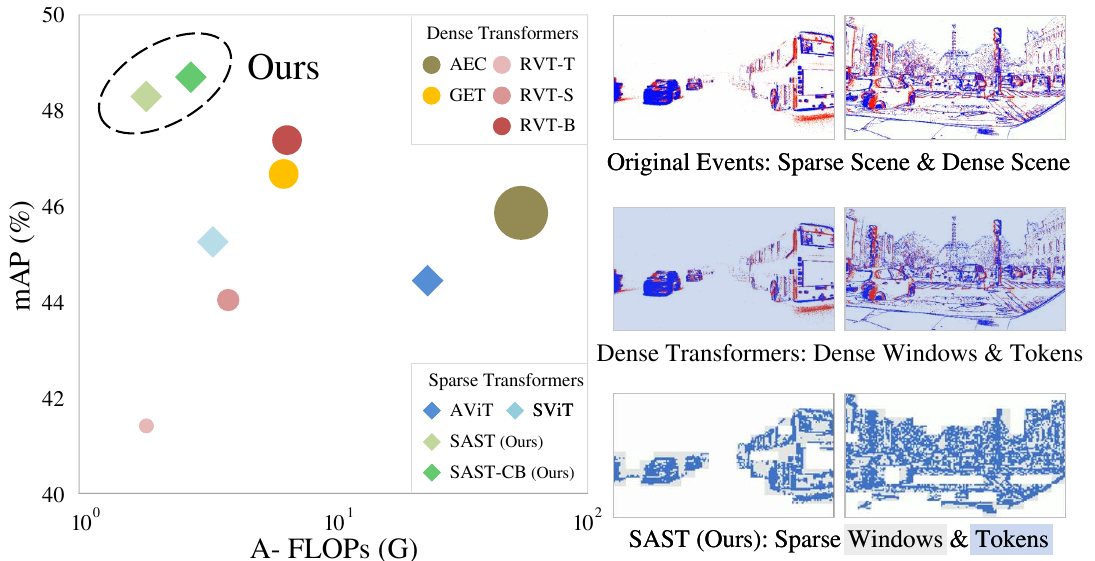}
 \vspace{-5mm}
\caption{\textbf{Detection performance vs. computational cost} on 1Mpx, with marker size indicating model size. SAST exhibits superiority by employing window-token co-sparsification and scene-specific sparsity optimization, maintaining low computation while delivering high performance across varying scenes.
}
 \vspace{-5mm}
\label{fig_teaser}
\end{figure}

\vspace{-3mm}
\section{Introduction}
\label{sec:intro}



Event cameras asynchronously capture the illumination changes of each pixel in a bionic way and possess several advantages, such as high temporal resolution (\textgreater 10K fps) and a wide dynamic range (\textgreater 120 dB) \cite{survey}. Unlike frame cameras that capture the whole scene at a fixed rate, event cameras exclusively record sparse event streams when objects are in motion, resulting in no events generated in static scenes. This distinctive trait enables event cameras to operate with low power consumption (\textless 10 mW) \cite{survey}. 

Building on these unique advantages of event cameras, event-based object detection excels in challenging motion and lighting conditions, and offers energy-efficient solutions in power-constrained environments \cite{SAM, ObjectDetection1, YOLO_DVS, RED, ASTM, AEC}.
However, as the raw events are asynchronous, traditional Image-based networks can not be directly applied to them. To overcome this problem, prior works often convert events into Image-like representations, such as event voxel \cite{VOXEL}, event histogram \cite{HIST}, and time surface \cite{HOTS, HATS} first.
Features are extracted from these representations using different neural network architectures, including CNN-based \cite{DNN1, DNN3, DNN4, DNN5}, SNN-based \cite{Wu2021TrainingSN, SNN-SSD, SNN3, GaborSNN, PLIF}, and GNN-based networks \cite{GNN1, GNN2}. Recently, many innovative works have demonstrated that Vision Transformers can achieve superior performance on event-based object detection tasks \cite{rvt, peng2023get, NeuralMemory}. However, the quadratic computational complexity of self-attention in Transformers hinders model scalability \cite{Tay_2022}, which challenges the balance between performance and efficiency for event-based object detection. 
The high power consumption of such dense operation also diminishes the low power consumption advantage of event cameras. 
Additionally, we observe that due to the high sparsity of events, a majority of the computational cost of the Transformer is unnecessary. Especially for high-resolution event streams, such as those in the 1Mpx \cite{RED} dataset, while higher resolution brings more detail, event sparsity can fluctuate by several hundred times. During self-attention computations, the blank regions without any events are also encoded as tokens and participate in calculations. Our goal is to reduce this unnecessary computational overhead. 

We revisit the mainstream Image-based sparse Transformers, which are also designed to reduce the computational overhead of self-attention. Most of them achieve token-level sparsification based on global self-attention \cite{AVIT, ADAVIT, kong2022spvit, Chen_2023_CVPR, chen2021chasing, Kim2021LearnedTP, IA-RED, rao2021dynamicvit}. 
Some studies among them achieve adaptive sparsification \cite{AVIT, ADAVIT}, but they are unable to choose the optimal sparsity for individual samples. 
What's more, despite computing much fewer tokens than the original ViT \cite{VIT}, they still require significantly more computational resources than window-based Transformers \cite{swin, swinv2, nest, maxvit} that leverage compute-efficient window self-attention. Consequently, the computational cost remains a barrier for high-resolution object detection tasks. SparseViT (SViT) \cite{chen2023sparsevit} proposes a sparse window-based Transformer architecture. However, it relies on a manually selected window pruning ratio, which may discard important windows containing objects. This results in poor robustness in certain scenarios, particularly when many objects smaller than the window scale are present.

In this work, we introduce Scene Adaptive Sparse Transformer (SAST) for event-based object detection. SAST achieves window-token co-sparsification in a window-based Transformer architecture. As shown in \cref{fig_teaser}, SAST greatly reduces computational overhead (A-FLOPs: Attention-related FLOPs, exclusive of the computations incurred by convolutional layers.), while simultaneously enhancing performance. Moreover, it leverages the scoring and selection modules, realizing scene-specific sparsity optimization which can adaptively choose the optimal sparsification strategy based on the complexity of different scenes. We also propose the Masked Sparse Window Self-Attention (MS-WSA), which efficiently performs self-attention on selected tokens with unequal window sizes and isolates all context leakage to achieve optimal performance. 
Our main contributions are summarized in the following.

\begin{itemize}
\setlength{\itemsep}{0pt}
\setlength{\parsep}{0pt}
\setlength{\parskip}{0pt}

\item We develop a highly efficient and powerful SAST for event-based object detection, which maintains a remarkable balance between performance and efficiency.

\item We propose innovative scoring and selection modules, which assess the importance of each window and token and perform co-sparsification on them.
\item We devise the MS-WSA, which efficiently performs window self-attention on selected tokens with unequal window sizes and avoids context leakage.
\item Experimental results on 1Mpx and Gen1 datasets demonstrate the superiority of SAST, surpassing all dense and sparse networks.
\end{itemize}

\section{Related Work}
\label{sec:relatedwork}

\noindent\textbf{Vision Transformers.}
The Vision Transformer (ViT) \cite{VIT} stands as a seminal work providing a vision backbone that applies self-attention to images. 
Based on that, many ViT variants have emerged to enhance performance and efficiency. For example, linear Transformers \cite{wu2022flowformer, Vyas2020FastTW, Performer, zhen2022cosformer} are explored for finding the approximation of self-attention. Other works \cite{maxvit, swin, t2tvit, TNT, localvit} introduce localized self-attention or hierarchical Transformer architectures. 

\noindent\textbf{Sparse Transformers.}
Sparse Transformers are proposed to improve the efficiency of ViTs by selectively computing self-attention on partial tokens. 
Most of them achieve token-level sparsification based on traditional ViT \cite{AVIT, ADAVIT, kong2022spvit, Chen_2023_CVPR, chen2021chasing, Kim2021LearnedTP, IA-RED, rao2021dynamicvit}. However, these sparsified networks still require significantly more computational resources than window-based Transformers \cite{maxvit, swin}. SViT \cite{chen2023sparsevit} proposes a window-level sparse Transformer based on Swin Transformer \cite{swin}. However, its manually selected window pruning ratio leads to the unwanted discarding of important windows. If the scene is dense or the objects are smaller than the window scale, performance will suffer severely.

\noindent\textbf{Event-Based Transformers.}
Transformer networks have achieved high performance across various event-based tasks, including but not limited to classification \cite{EvT}, object detection \cite{rvt, peng2023get}, and semantic segmentation \cite{NeuralMemory}. However, due to its high computational complexity, many early attempts only incorporate self-attention operations on feature maps derived from CNN backbones \cite{Kim_2023_CVPR, TransformerVideoReconstruction, TransformerOpticalFlow, TransformerDomainAdaptation}. Recently, alternative approaches \cite{Transformer1, Transformer2, EvT, rvt, peng2023get} using efficient window Transformers aim to extract features from event representations directly \cite{VOXEL, HIST, HOTS, HATS}. 
However, a substantial computational burden of self-attention still exists for high-resolution tasks \cite{NeuralMemory, rvt}, which limits the model's scalability and performance \cite{Tay_2022}.

\begin{figure*}[t]
\centering
\includegraphics[width=2\columnwidth]{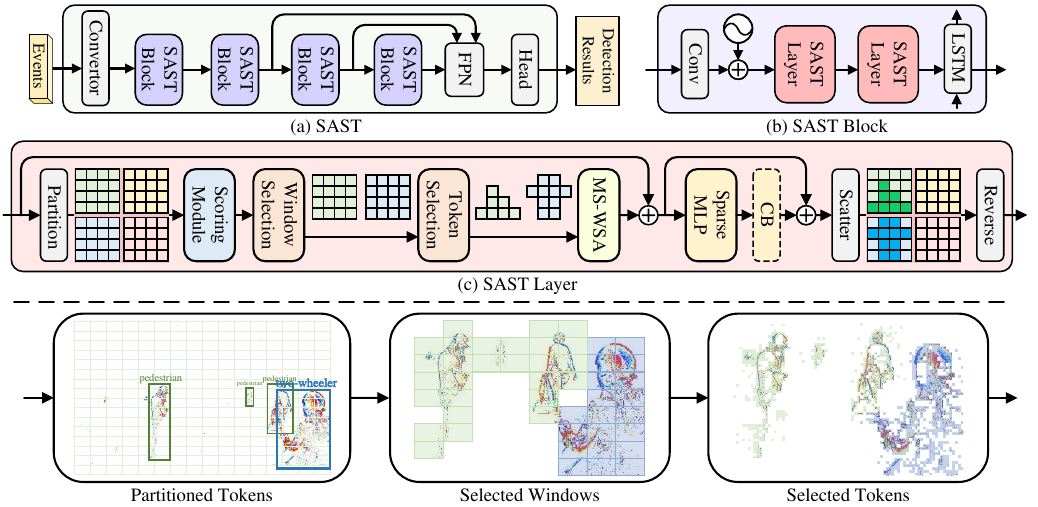}
 \vspace{-2mm}
\caption{ 
\textbf{(a) The hierarchical architecture of SAST.} Four SAST blocks extract multi-scale features from sparse tokens transformed from events. \textbf{(b) The architecture of SAST block}, which contains two successive SAST layers.
\textbf{(c) The architecture of SAST layer.} In an SAST layer, tokens are partitioned into windows and scored by the scoring module first. The selection module selects important windows and tokens. Then, the selected tokens within selected windows are sequentially processed through MS-WSA, a sparse MLP layer, and an optional CB operation. Finally, processed tokens are scattered back and reversed from windows. Norm layers are omitted for simplification.}
 \vspace{-2mm}
\label{fig_Layer}
\end{figure*}

\noindent\textbf{Event-Based Object Detection}
The early efforts in event-based object detection involve converting event streams into images and feeding them into existing Image-based detectors \cite{image2,mnistimage}. Significant strides were made with the advent of benchmark large-scale datasets 1Mpx \cite{RED} and Gen1 \cite{gen1}. They paved the way for the development of innovative networks such as RED \cite{RED} and ASTMNet \cite{ASTM}, which introduce memory mechanisms to fully exploit the spatiotemporal information within event data. Some approaches based on SNNs and GNNs were also proposed \cite{SNN-SSD, GNN2}, but their performance still lags behind CNN-based methods. Recently, remarkable performance has been achieved by RVT \cite{rvt}, HMNet \cite{NeuralMemory}, and GET \cite{peng2023get} using Transformer networks. However, these networks still involve significant redundancy in self-attention computations of sparse events.

\section{Method}
\subsection{Main Architecture}
The overall architecture of SAST is shown in \cref{fig_Layer}(a).
The asynchronous events are initially converted into event voxel representations \cite{VOXEL}. These event voxels are fed into four hierarchical SAST blocks illustrated in \cref{fig_Layer}(b) to extract multi-scale features. Within the first block, a convolutional layer, as in ViT \cite{VIT}, transforms the event voxel into tokens. Subsequently, the sinusoidal positional encoding is added to the tokens. Two successive SAST layers extract spatial features from these tokens. At the end of the SAST block, an LSTM layer propagates temporal information, with its output sent to subsequent layers. In the following blocks, tokens undergo a convolutional layer first to reduce the spatial resolution, the remaining process being the same as in the first block. The extracted features from the second, third, and fourth blocks are relayed to the Feature Pyramid Network (FPN). The FPN then forwards the processed features to the detection head, yielding the final detection results.

\begin{figure*}[t]
\centering
\includegraphics[width=2\columnwidth]{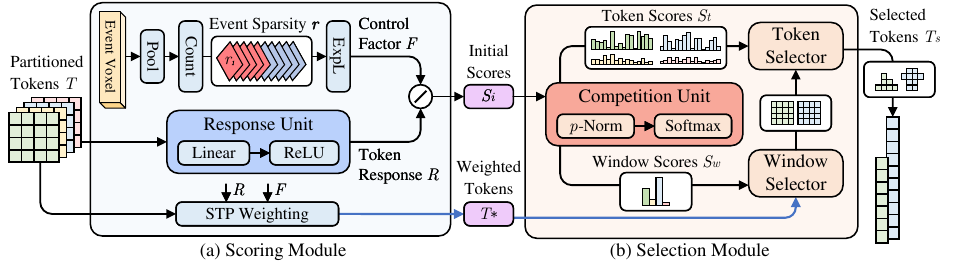}
 \vspace{-2mm}
\caption{\textbf{The architecture of scoring and selection modules.} (a) The scoring module scores each window and token. The scoring process is regulated based on event sparsity, with windows and tokens competing to limit selections. (b) The selection module uses two filters to select important windows and tokens sequentially based on their scores.
 }
 \vspace{-2mm}
\label{fig_Scoring}
\end{figure*}
\subsection{SAST Layer}
As shown in \cref{fig_Layer}(c), in an SAST layer, all the tokens are partitioned into windows first. After that, a scoring module scores the partitioned tokens, determining their importance. The selection module selects important windows and tokens inside windows sequentially based on these scores. The token selection results are shared between SAST layers within the same block. Next, the Masked Sparse Window Self-Attention (MS-WSA) is performed on selected tokens within selected windows. The attention-enhanced tokens are sequentially processed by a sparse MLP layer and an optional context broadcasting (CB) operation \cite{hyeon2022scratching}.
The sparse MLP layer indicates that sparse connections are established exclusively on selected tokens, thereby benefiting from sparsification and reducing computational load. 
Finally, the processed tokens are scattered back and reversed from windows to their original shape. 

The SAST layer achieves sparsification on both windows and tokens, ensuring efficiency and preventing important windows containing objects from being discarded. Its scene-aware adaptability also guides the network to focus more on important areas, resulting in better performance.
What's more, the computational cost of SAST is fully dynamic across different scenes, while maintaining an overall low level.
The SAST layer is compatible with different Transformer architectures, which means it can be plugged as a universal layer in any hierarchical Transformer. We then detail the three main components of the SAST layer: the scoring module, the selection module, and MS-WSA.

\subsection{Scoring Module}
As illustrated in \cref{fig_Scoring}(a), the scoring module is designed to score each token, determining its importance. Unlike other sparse Transformers obtaining scores directly from the token value or attention map, SAST has a controllable and learned scoring module for reasonable scoring.

Initially, the partitioned tokens $T$ are sent to a response unit to obtain their token responses $R$. The response unit consists of a linear layer followed by a ReLU layer. On a parallel branch, due to multi-scale design, the original event voxel is downsampled by a pooling layer first to match the receptive field of tokens. Event sparsity $\textbf{r}$ is obtained by calculating and concatenating $B$ non-zero ratios $(r_1, r_2, ..., r_B)$ of different voxel bins. Each ratio reflects the sparsity of a specific subset of events with unique time ranges and polarities.
Then, the event sparsity $\textbf{r}$ is projected to the same dimension as the tokens through the Exponential Linear layer (ExpL), resulting in the control factor $F$. Based on the definition of the token response $R$ and the control factor $F$, the initial score $S_i$ is defined as:
\begin{equation}
S_i = a \cdot \frac{R}{F} = a \cdot \frac{\text{ReLU}(W_R \cdot T + b_R)}{\exp(W_F) \cdot \textbf{r}},
\label{eq:initialscores}
\end{equation}
where $W_R$ and $b_R$ represent the weight and bias of the linear layer in the response unit, while 
$W_F$ represents the weight of the ExpL. The function 
$\exp(\cdot)$ denotes the exponential operation, which is involved to ensure that the control factor $F$ is positive. $a$ is a hyper-parameter controlling the absolute sparsity level of SAST. 
According to our observation, tokens corresponding to more important objects tend to have higher initial scores. 

To make the response unit and the ExpL learnable, we apply Spatial-Temporal-Polar (STP) weighting to the partitioned tokens $T$. A Sigmoid layer transforms the token response $R$ and the control factor $F$ into spatial weight $W_s$ and temporal-polar weight $W_{tp}$. The weighted tokens $T\ast$ are produced through the product of these weights, as described by the equation:
\begin{equation}
T\ast = W_{tp}\cdot W_s \cdot T=\text{Sigmoid}(R) \cdot \text{Sigmoid}(F) \cdot T.
\label{eq:weighting}
\end{equation}

The STP weighting process allows the model to emphasize tokens based on their spatial and temporal-polar context, aligning the network's sparse processing with the most salient features across both domains.

\subsection{Selection Module}
As illustrated in \cref{fig_Scoring}(b), the selection module is designed to select important tokens and windows.
However, the initial scores derived from normalized tokens show insufficient contrast for effective differentiation. Therefore, the competition unit is designed to intensify the competition among the $S_i$. It first calculates the $p$-norm of the $S_i$, obtaining the normalized score for each token and window, respectively. 
The choice of $p$ determines the exponential relationship between the normalized score and event sparsity. Softmax operation is further applied to amplify the disparities between normalized scores, making the dominance of some values more pronounced. The intensified scores are calculated using the equations:
\begin{equation}
\begin{split}
&S_{t} = \text{Softmax}(||S_{i}||_p^\text{c}) \\
&S_{w} = \text{Softmax}(||S_{i}||_p^\text{c, w}/N_{t})
\end{split}
\label{eq:softmax}
\end{equation}
where $S_{t}$ and $S_{w}$ are the token scores and window scores, respectively. $||.||_p^\text{c}$ denotes computing $p$-norm along the channel dimension, and $||.||_p^\text{c,w}$ denotes computing $p$-norm along both the channel and window dimensions. The division by $N_{t}$ serves to scale the normalized window scores by the number of tokens in a window.

If the scene is sparse, fewer scores dominate, making it easier to select less important windows and tokens. Conversely, in dense scenes, the score distribution is relatively uniform, preserving more windows and tokens to prevent the loss of important object information. 
Then, the selection module specifies two thresholds $\mu_{t}$ and $\mu_{w}$ for the selection process, defined as:
\vspace{-0.5mm}
\begin{equation}
\mu_{t} = \frac{b}{N_{t}},\quad\mu_{w} = \frac{b}{N_{w}},
\label{eq:criterion}
\end{equation}
where $N_{t}$ is the number of tokens in a window, and $N_{w}$ is the number of windows. $b$ is a parameter in the range [0.9,1] that controls the strictness of selection. Therefore, $\mu_{t}$ and $\mu_{w}$ correspond to values slightly smaller than the mean of token and window scores. Scores exceeding the threshold reflect their significance, ensuring that the selected proportion aligns with scene complexity.


In \cref{fig_Scoring}(b), weighted tokens $T\ast$ undergo an initial selection by a window selector that retains tokens from windows with scores $S_{w}$ exceeding $\mu_{w}$. Then, a token selector further refines this selection, keeping tokens with scores $S_{t}$ surpass $\mu_{t}$. The final selected tokens are represented as $T_s$.

\begin{figure}[t]
\centering
\includegraphics[width=\columnwidth]{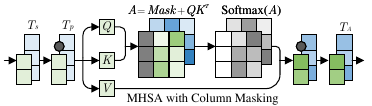}
 \vspace{-5mm}
\caption{\textbf{Masked Sparse Window Self-Attention (MS-WSA).} An illustration of MS-WSA with window size equal to $2\times 2$. Three and two tokens are selected from two windows. Due to the column masking, the padded token marked as a grey dot does not contribute to the attention map.
 }
 \vspace{-3mm}
\label{fig_WSA}
\end{figure}
\subsection{Masked Sparse Window Self-Attention}

Traditional WSA is designed for equal-sized windows, conducting parallel matrix multiplication on them. As a result, it cannot be applied to selected tokens with unequal window sizes. 
Therefore, we design MS-WSA to efficiently perform self-attention on selected tokens, integrating correlations among them and isolating context leakages. 

As shown in \cref{fig_WSA}, MS-WSA first pads the selected tokens $T_s$ within each window to the same length. The padding operation involves selecting a minimal number of tokens from the filtered-out tokens. The padded tokens are represented as $T_p$. 
Afterward, we apply multi-head self-attention (MHSA) \cite{Attentionis} in parallel to selected windows containing padded tokens. However, the tokens used for padding, originally unselected, also contribute to the attention map calculation. Moreover, the padding number is determined by the largest window in the batch. These uncertainties lead to contextual leakage both among tokens and across different batches. To isolate these context leakages, MS-WSA includes an additional masking operation when calculating the attention map. The overall MS-WSA process can be described as:
\vspace{-0.5mm}
\begin{equation}
\begin{split}
&T_p = \text{Pad}(T_s) \\ 
&Q,K,V= T_pW^Q, T_pW^K, T_pW^V\\
&T_A=\text{UnPad}(\text{Softmax}(Mask+QK^{T})V),
\end{split}
\end{equation}
where $\text{Pad}(\cdot)$ indicates the padding operation. $W^Q$, $W^K$, and $W^V$ are linear weights to transform padded tokens into query $Q$, key $K$, and value $V$. The $Mask$ matches the attention map $A=QK^{T}$ in size and contains large negative values in the columns corresponding to the tokens used for padding, while other values are set to zero. The masking operation cuts off the influence of unselected tokens in the Softmax process. After a matrix multiplication operation between the post Softmax attention map and $V$, the tokens at padded locations are removed by the $\text{UnPad}(\cdot)$ operation, resulting in attention-enhanced tokens $T_A$.

After performing MS-WSA, the attention-enhanced tokens $T_A$ sequentially undergo a sparse MLP layer and an optional context broadcasting (CB) operation \cite{hyeon2022scratching}. The CB operation can be expressed as:
\begin{equation}
\text{CB}(\widetilde{T}(n)) = \frac{1}{2}\widetilde{T}(n) + \frac{1}{2N} \sum_{n=1}^N \widetilde{T}(n) \  \text{for}\ n=1,\dots,N\textbf{},
\end{equation}
where $\widetilde{T}(n)$ represents the $n_{th}$ output token of the sparse MLP layer. $N$ is the total number of selected tokens.
The CB operation broadcasts information among selected tokens, enriching their information density to ensure that they will be consistently selected in subsequent layers. It does not incur additional computation but leads to a looser sparsification preference for SAST. 
Finally, the processed tokens are scattered back to the original partitioned tokens $T$ and reversed from windows to their original shape. 
\section{Experiments}

\begin{table*}[t]
  \centering
  \tabcolsep=0.098cm
  \begin{adjustbox}{width=2\columnwidth,center}
    \begin{tabular*}{\hsize}{@{}@{\extracolsep{\fill}}lc|cc|cc|c@{}}
      \multicolumn{2}{c}{} 
      & \multicolumn{2}{c}{1Mpx} 
      & \multicolumn{2}{c}{Gen1} 
      &  \\
      \cline{1-7}
      Methods
      & Backbone
      & mAP (\%) & FLOPs (G) 
      & mAP (\%) & FLOPs (G) 
      & Params (M)
      \\
      \hline
      SAM \cite{SAM} & ResNet50 \cite{resnet}
      & 23.9 & 19.0 
      & 35.5 & 6.0 
      & \textgreater 20 \\
      YOLOv3\_DVS \cite{YOLO_DVS} & Darknet-53 \cite{yolov3}
      & 34.6 & 34.8 
      & 31.2 & 11.1 
      & \textgreater 60 \\      
      RED \cite{RED} & ResNet50 \cite{resnet}
      & 43.0 & 19.0 
      & 40.0 & 6.0 
      & 24.1 \\

      ASTMNet \cite{ASTM} & VGG16 \cite{vgg}
      & 48.3 & 75.7 
      & 46.7 & 29.3 
      & \textgreater 100 \\

      AEC \cite{AEC} & CSP-Darknet-53 \cite{yolov5}
      & 48.4 & 58.2 
      & 47.0 & 20.9 
      & 46.5 \\

      \hline
      AEC \cite{AEC} & Deformable-DETR \cite{DDETR}
      & 45.9 & \textgreater 50 
      & 44.5 & \textgreater 20 
      & 40.0 \\
      GET \cite{peng2023get}  & GET \cite{peng2023get}
      & 48.4 & 10.6\ (6.3) 
      & 47.9 & 3.6\ (2.2) 
      & 21.9 \\
      RVT \cite{rvt} & Swin Transformer\cite{swin}
      & 46.7 & 10.4\ (6.6) 
      & 44.4 & 3.6\ (2.3) 
      & 18.5 \\
      RVT \cite{rvt} & ConvNeXt \cite{convnext}
      & 45.5 & 10.4\ (6.6) 
      & 42.3 & 3.6\ (2.3) 
      & 18.7 \\
      RVT \cite{rvt} (baseline) & MaxViT \cite{maxvit}
      & 47.4 & 10.3\ (6.5) 
      & 47.2 & 3.5\ (2.2) 
      & 18.5 \\

      \hline
      Ours & SAST
      & 48.3 (\textcolor{ForestGreen}{+0.9})& \ \ \textbf{5.6}\ (\textbf{1.8},\ \textcolor{ForestGreen}{-72$\%$}) 
      & 47.9 (\textcolor{ForestGreen}{+0.7})& \textbf{2.1}\ (\textbf{0.8},\ \textcolor{ForestGreen}{-64$\%$}) 
      & 18.9 \\
      Ours & SAST-CB
      & \textbf{48.7} (\textcolor{ForestGreen}{+1.3})& \ \ 6.4\ (2.6,\ \textcolor{ForestGreen}{-60$\%$}) 
      & \textbf{48.2} (\textcolor{ForestGreen}{+1.0})& 2.4\ (1.1,\ \textcolor{ForestGreen}{-50$\%$}) 
      & 18.9 \\

      \hline
    \end{tabular*}
  \end{adjustbox}
 \vspace{-2mm}
  \caption{
  Detection performance compared with state-of-the-art methods on 1Mpx and Gen1.
  The reported FLOPs belongs to the backbone. Values in brackets ($\cdot$) indicate the A-FLOPs, excluding operations from convolutional layers, and is precisely the value we aim to reduce.}
 \vspace{-3mm}
  \label{tab_SOTA}%
\end{table*}

Initially, we outline the experimental setup, detailing the datasets used, evaluation metrics, and various implementation specifics. Next, we compare SAST with other state-of-the-art works on two large-scale event-based object detection datasets. We also compare SAST with baseline variants and other sparsification methods. Finally, we offer insights through ablation studies, visualizations, and statistical results to thoroughly examine the effectiveness of our proposed sparsification approach.
\subsection{Experimental Setup}
\noindent\textbf{Datasets.}
The 1Mpx dataset, commonly used for event-based object detection, consists of 14.65 hours of events, with a large resolution of $1280\times720$ pixels. It includes 7 labeled object classes. We follow previous works \cite{ASTM, RED, rvt}, utilizing 3 classes: car, pedestrian, and two-wheeler for performance comparison. This dataset has a labeling frequency of 60 Hz and contains over 25M bounding boxes.
The Gen1 dataset comprises 39 hours of events. It has a smaller resolution of $304\times240$ pixels and contains 2 object classes. The labeling frequency for this dataset is 20 Hz.

\noindent\textbf{Metrics.}
We use the COCO mean average precision (mAP) \cite{COCO} as the main metric. 
To measure computational complexity, we compute the average FLOPs (Floating Point Operations Per Second) over the first 1000 samples in the test sets of 1Mpx and Gen1, referring to Sparse DETR \cite{roh2022sparse} that calculates the FLOPs for the first 100 samples in the MS COCO dataset. 
Additionally, we report the average Attention-related FLOPs (A-FLOPs), which excludes the computations incurred by convolutional layers during calculation. The model's size is measured through its parameter count.
We also compare the inference time (runtime) with baseline variants and different sparsification methods.

\noindent\textbf{Implementation Details.}
We adopt the predefined train, validation, and test splits of 1Mpx and Gen1 datasets. The accumulation time of each sample is chosen as 50 ms. We also follow the dataset preprocessing methods of previous works \cite{ASTM, rvt, RED, peng2023get}, such as removing misleading small bounding boxes and downsampling the events in the 1Mpx dataset into $640\times360$ for a fair comparison. We choose RVT-B (RVT) \cite{rvt} as the baseline and apply different sparsification methods to its MaxViT backbone, while maintaining consistency in the design of other layers (FPN, Head, and LSTM). The partition strategy is the same as RVT, dividing tokens into windows and grids (a window type) in two successive SAST layers. The augmentation strategies include zoom-in, zoom-out, and horizontal flipping. We utilize eight NVIDIA TITAN Xp GPUs for training. For the testing phase, we only use one NVIDIA TITAN Xp GPU.

\subsection{Experimental Results}
\noindent\textbf{Comparison with SOTAs.}
We evaluate SAST on the 1Mpx and Gen1 datasets. The results compared with state-of-the-art works are reported in \cref{tab_SOTA}. 
To facilitate distinction, we define the model as SAST-CB when the optional CB operation is utilized.

On the 1Mpx dataset, our methods outperform all Transformer-based networks, including AEC (Deformable-DETR as the backbone), GET \cite{peng2023get}, and RVT \cite{rvt} using different backbones \cite{swin, maxvit, convnext}. SAST achieves a $48.3\%$ mAP with only $28\%$ of the A-FLOPs of RVT. However, in comparison to all works involving pure convolutional networks, SAST falls slightly below AEC (CSP-Darknet-53 as the backbone), which utilizes a more complex event representation, introduces additional data augmentations, and has 9.4 times more FLOPs. SAST-CB achieves a superior result of $48.7\%$ mAP, surpassing all other state-of-the-art networks with merely $11\%$ of AEC's FLOPs and $40\%$ of the A-FLOPs compared to RVT.

On the Gen1 dataset, both SAST and SAST-CB are superior to all other methods, achieving $47.9\%$ and $48.2\%$ mAP, which is $0.7\%$ and $1.0\%$ higher than RVT. Moreover, the FLOPs and A-FLOPs of SAST are only $60\%$ and $36\%$ of RVT, respectively. In comparison to the pure convolutional network AEC (CSP-Darknet-53 as the backbone), SAST and SAST-CB achieve $0.9\%$ and $1.2\%$ performance gain with just $10\%$ and $11\%$ of the FLOPs.
\begin{table}[t]
  \centering
  \tabcolsep=0.098cm
  \begin{adjustbox}{width=\columnwidth,center}
    \begin{tabular*}{\hsize}{@{}@{\extracolsep{\fill}}l|cc|c@{}}

      \cline{1-4}
      Methods
      & mAP (\%) & A-FLOPs (G) 
      & runtime$^\ast$ (ms)
      \\
      \hline

      RVT-T \cite{rvt}
      & 41.5 & \textbf{1.8} 
      & 14.5 \\
      RVT-S \cite{rvt}
      & 44.1 & 3.8 
      & 15.3 \\
      RVT \cite{rvt} (baseline)
      & 47.4 & 6.5 
      & 16.0 \\
      AViT \cite{ADAVIT}
      & 44.5 & 23.4 
      & 37.0 \\
      SViT \cite{chen2023sparsevit}
      & 45.3 & 3.3 
      & 14.3 \\
      SparseTT \cite{sparsett}
      & 47.6 & 6.5 
      & 16.1 \\
      SAST (Ours)
      & \textbf{48.3} & \textbf{1.8} 
      & 19.7 \\
      \hline
    \end{tabular*}
  \end{adjustbox}
 \vspace{-2mm}
  \caption{Detection performance on 1Mpx using RVT variants and different sparsification methods. SAST achieves optimal performance with the least computational expense. $\ast$: All the runtime is tested on one NVIDIA TITAN Xp GPU.}
 \vspace{-5mm}
  \label{tab_sparse}%
\end{table}
\begin{figure}[t]
\centering
\includegraphics[width=\columnwidth]{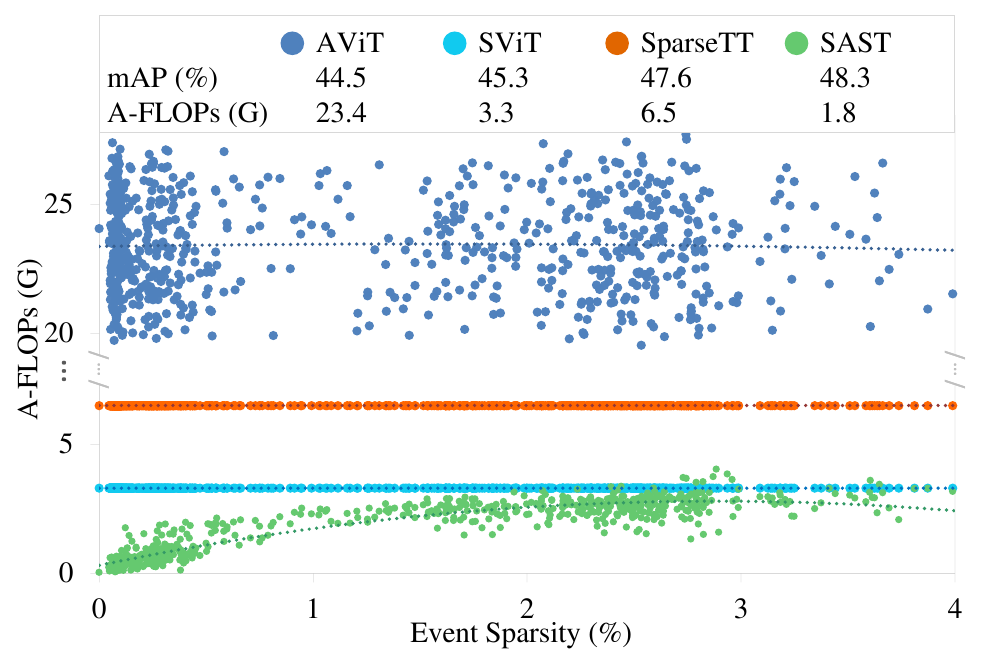}
 \vspace{-6mm}
\caption{
The computational complexity of different networks under scenes with varying event sparsity in 1Mpx. Each point represents an event sample. SAST adaptively adjusts its sparsity based on the scene complexity.}
 \vspace{-4mm}
\label{fig_scatter}
\end{figure}

\noindent\textbf{Comparison with RVT variants \& sparse Transformers.}
As shown in \cref{tab_sparse}, we further compare SAST with RVT variants and sparse Transformers on the 1Mpx dataset.

RVT variants reduce A-FLOPs significantly by scaling down the model size, at the cost of substantial performance decline.
AViT \cite{AVIT} implements token-level sparsification based on ViT. However, the computational complexity of its global self-attention after sparsification remains high. Its A-FLOPs exceeds $260\%$ of the baseline, with a performance decrease of $2.9\%$. SViT \cite{chen2023sparsevit}, which builds upon window-based Swin-Transformer \cite{swin}, achieves window-level sparsification at a manually set window pruning ratio (set to $50\%$ as in the paper), lacking adaptability. Although its A-FLOPs is only $50.8\%$ of RVT's, due to the removal of a fixed number of windows in all scenarios, its performance dropped by $2.1\%$. SparseTT \cite{sparsett} achieves a $0.2\%$ performance gain by sparsifying the attention map. However, such mask-based sparsification does not actually reduce computational costs. Our proposed SAST, thanks to its outstanding scene-adaptive sparsification, achieves the highest mAP of $48.3\%$, with only $7.7\%$ A-FLOPs of AViT and $54.5\%$ A-FLOPs of SViT.

In \cref{fig_scatter}, we illustrate the computational complexity (A-FLOPs) for the first 1000 samples in the 1Mpx dataset. 
It can be observed that both RVT and SViT maintain constant A-FLOPs regardless of changes in the scene. AViT shows some level of adaptability, which remains evenly distributed around its average A-FLOPs. Our proposed SAST shows remarkable scene-aware adaptability. It has very low A-FLOPs in sparse scenes while dynamically optimizing the sparsity level to maintain high performance in dense scenes.

\subsection{Ablation Studies}

To analyze the proposed sparse Transformer SAST, we conduct a series of ablative experiments on the 1Mpx dataset. 

\begin{table}[t]
  \centering
  \tabcolsep=0.098cm
  \begin{adjustbox}{width=\columnwidth,center}
    \begin{tabular*}{\hsize}{@{}@{\extracolsep{\fill}}lc|cc@{}}
      \cline{1-4}
      Scoring Methods&
      Source 
      & mAP (\%) & Params (M) 
      \\
      \hline
      L2 Activation
      & SparseViT \cite{chen2023sparsevit}
      & 45.1
      & 18.6 \\
      Attention Mask
      & SparseTT \cite{sparsett}
      & 45.8
      & 18.6 \\
      Head Scores
      & AS-ViT \cite{AVIT}
      & 46.8
      & 18.7 \\
      Head Importance
      & SPViT \cite{kong2022spvit}
      & 46.5
      & 18.9 \\
      Scoring Module
      & SAST (Ours)
      & \textbf{48.3}
      & 18.9 \\
      \hline
    \end{tabular*}
  \end{adjustbox}
  \vspace{-2mm}
  \caption{
  Detection performance on 1Mpx by using different scoring methods. The scoring module of SAST surpasses other established scoring methods in sparse Transformers. }
  \vspace{-2mm}
  \label{tab_ablscoring}%
\end{table}
\begin{table}[t]
  \centering
  \tabcolsep=0.098cm
  \begin{adjustbox}{width=\columnwidth,center}
    \begin{tabular*}{\hsize}{@{}@{\extracolsep{\fill}}l|cc|c@{}}

      \cline{1-4}
      Selection Target
      & mAP (\%) & A-FLOPs (G) 
      & runtime (ms)
      \\
      \hline
      Window
      & 46.3 & 3.7 
      & 20.1 \\
      Token
      & 47.7 & 3.0
      & 21.7 \\
      Window \& Token
      & \textbf{48.3} & \textbf{1.8}
      & 19.7 \\
      \hline
    \end{tabular*}
  \end{adjustbox}
  \vspace{-2mm}
  \caption{
  Detection performance on 1Mpx across different selection targets.
  Selection on both windows and tokens forces the network to focus on the most salient and crucial features, resulting in improved performance.}
  \vspace{-2mm}
  \label{tab_ablselecting}%
\end{table}
\begin{table}[t]
  \centering
  \tabcolsep=0.088cm
  \begin{adjustbox}{width=\columnwidth,center}
    \begin{tabular*}{\hsize}{@{}@{\extracolsep{\fill}}l|c|cc@{}}

      \cline{1-4}
      Methods
      & Context Leakage
      & mAP (\%) & A-FLOPs (G)
      
      \\
      \hline
      SA
      & Yes (batch) 
      & 40.7 & 16.5\\
      S-SA
      & Yes (token)
      & 44.2 & 20.5\\
      MS-SA
      & No Leakage
      & 46.6 & 21.2\\
      WSA
      & N/A 
      & N/A & N/A\\
      S-WSA
      & Yes (token)
      & 46.2 & 2.0\\
      MS-WSA
      & No Leakage
      & \textbf{48.3} & \textbf{1.8}\\
      \hline
    \end{tabular*}
  \end{adjustbox}
  \vspace{-2mm}
  \caption{Detection performance on 1Mpx using different self-attention methods. MS-WSA efficiently performs WSA on selected tokens with unequal window sizes, optimizing performance by preventing context leakage between batches and tokens.}
  \vspace{-4mm}
  \label{tab_ablweight}%
\end{table}
\begin{figure*}[t]
\centering
\includegraphics[width=2\columnwidth]{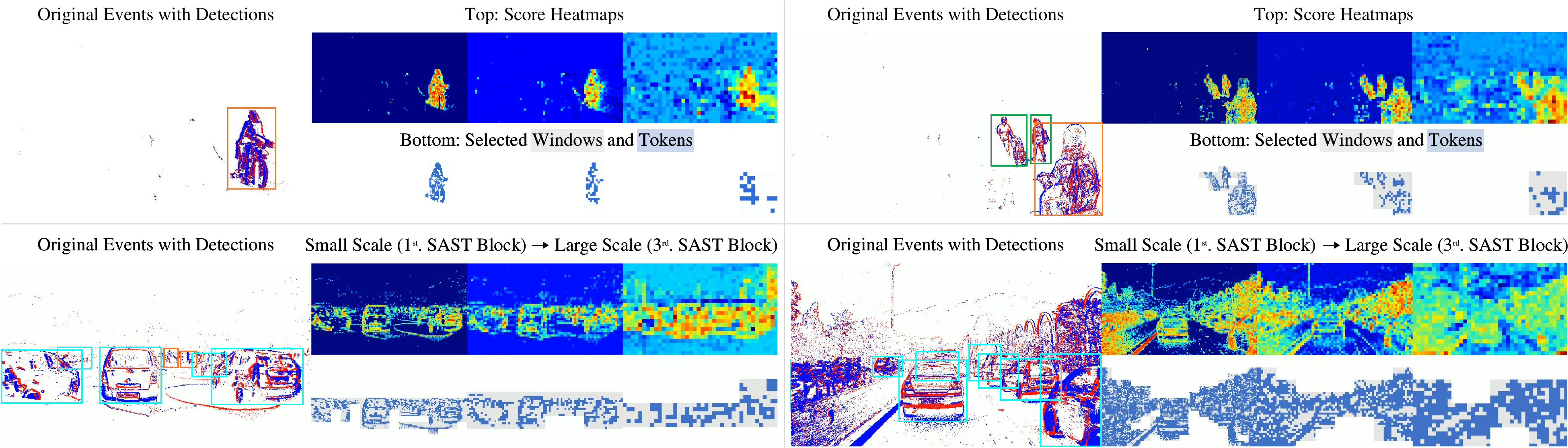}
 \vspace{-2mm}
\caption{Visualizations of original events, score heatmaps, and selection results under four scenes in 1Mpx. 
As the network progresses through subsequent SAST blocks, featuring multiple downsampling stages, the scale (receptive field) of tokens expands.
}
 \vspace{-3mm}
\label{fig_visualizations}
\end{figure*}
\begin{figure}[t]
\centering
\includegraphics[width=\columnwidth]{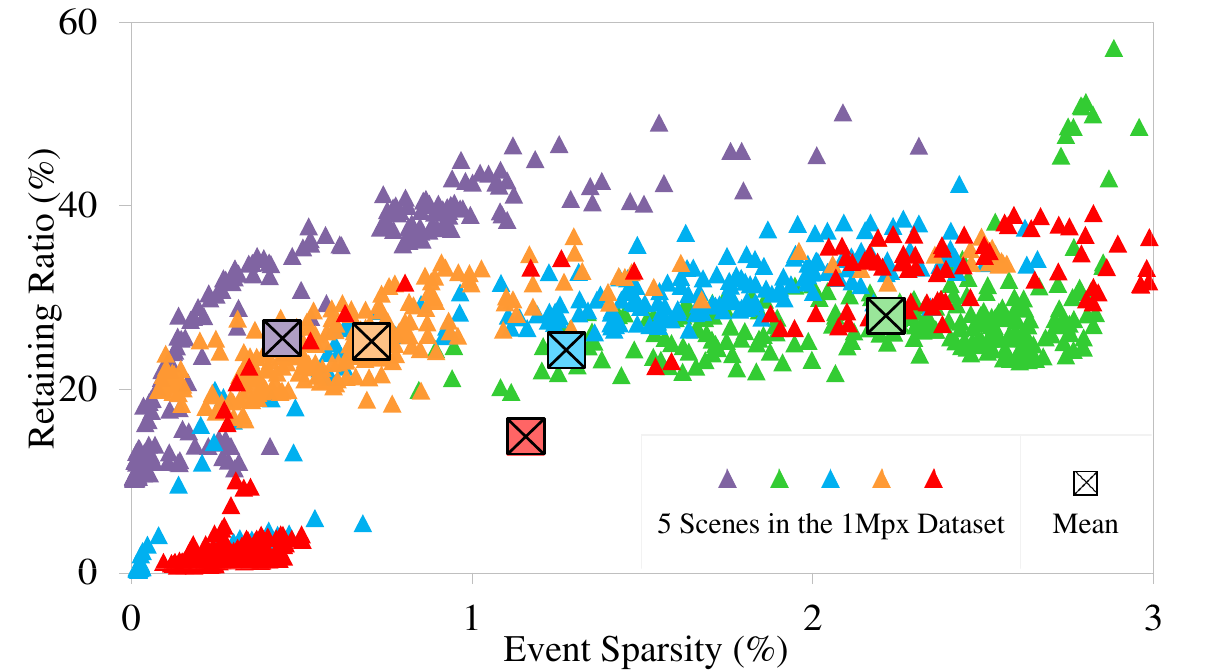}
 \vspace{-6mm}
\caption{
The averaged token retaining ratio of SAST across five scenes containing different objects in 1Mpx. Each $\triangle$ corresponds to an event sample. The evident fluctuations and clusters demonstrate the model's scene-aware adaptability.}
 \vspace{-3mm}
\label{fig_keep}
\end{figure}
\noindent\textbf{Scoring Method.}
We interchanged our scoring module with various established methods from existing sparse Transformers while maintaining consistency in other architectural designs. The scoring module of SAST, as highlighted by the superior mAP of $48.3\%$ in \cref{tab_ablscoring}, outperforms other scoring methods. Leveraging STP weighting across both spatial and temporal-polar domains, SAST provides a more effective and context-aware approach for evaluating token importance.

\noindent\textbf{Selection Method.}
We compare different selection methods by training two SAST variants that only select windows and tokens, respectively. As shown in \cref{tab_ablselecting}, applying selection to both windows and tokens achieves the highest mAP of $48.3\%$ and the lowest A-FLOPs of $1.8G$. This strategy limits the amount of information the model can use, which forces the model to concentrate on the most salient and crucial features. Consequently, it can lead to a more efficient representation where the model learns to compress token information more densely.




\noindent\textbf{Self-Attention Method.}
We compare different self-attention methods by applying them to selected tokens with varying window sizes. It can be seen from \cref{tab_ablweight} that standard self-attention (SA) needs to process all tokens together, leading to context leakage across different batches and adversely affecting performance. Sparse Self-Attention (S-SA) employs padding to isolate tokens from different batches, but still results in context leakage between selected and padded tokens. Masked Sparse Self-Attention (MS-SA) further includes masking operation, which prevents both types of context leakage. However, as global self-attention, it still has high A-FLOPs. Window Self-Attention (WSA) can not be applied to windows of unequal size; after padding, Sparse Window Self-Attention (S-WSA) becomes usable but introduces context leakage between tokens, resulting in performance degradation. Only by applying MS-WSA, which isolates all context leakage and is fully parallel, is it possible to achieve optimal performance with the least computational complexity.




\subsection{Adaptability Analysis}
\noindent\textbf{Visualizations.}
We train SAST on the 1Mpx dataset and infer it on four typical scenes in the test set for visualizations. The event sparsity and scene complexity increase across the four scenes progressively. In \cref{fig_visualizations}, we provide visualizations of the original events, score heatmaps, and the selection results of windows and tokens. For the score heatmaps and selection results, figures from left to right depict the transition from first to third SAST blocks. In a later block, the number of tokens decreases, and their scale (receptive field) enlarges. These intuitive visualizations allow us to assess the model's adaptability more comprehensively.
From the visualizations of score heatmaps, we observe that the network demonstrates its scene-aware adaptability to assign important tokens higher scores. From the visualizations of window and token selection results, in complex scenes, the network proactively reduces the sparsity level to retain more important tokens.

\noindent\textbf{Token Retaining Ratio.}
The scatter plot in \cref{fig_keep} illustrates the relationship between the token retaining ratio and event sparsity across five scenes in the 1Mpx dataset.
The general trend in the figure shows that SAST retains more tokens as event sparsity increases, demonstrating its ability to optimize the sparsity level based on the scene complexity.
Furthermore, there are significant fluctuations in token retaining ratios at equivalent event sparsity. This reveals that SAST's optimization is not only based on the event sparsity but also on the distinct characteristics of each scene, such as the classes, numbers, and sizes of objects within the scene. This is also evidenced by the pronounced clustering of samples from distinct scenes, indicating that SAST can perceive the underlying patterns of different scenes, adaptively tuning its sparsity strategy accordingly.
In essence, the scatter plot demonstrates that SAST can dynamically tailor its sparsity level to the specific demands of each scene, thereby exhibiting scene-aware adaptability.

\section{Conclusion}
In this paper, we provide a novel vision Transformer for event-based object detection, called Scene Adaptive Sparse Transformer (SAST). SAST's adaptive sparsification mechanism enables window-token co-sparsification, significantly reducing the computational overhead. Utilizing the novel
scoring module, selection module, and MS-WSA, SAST showcases scene-aware adaptability, dynamically optimizing its sparsity across different scenes for high performance. Our results confirm the effectiveness of SAST, achieving state-of-the-art mAP on the 1Mpx and Gen1 datasets while maintaining remarkable computational efficiency. 

\section{Acknowledgement}

This work was in part supported by the National Natural Science Foundation of China (NSFC) under grants 62032006 and 62021001.

{
    \small
    \bibliographystyle{ieeenat_fullname}
    \bibliography{main}
}
\clearpage

\setcounter{section}{0}
\setcounter{page}{1}
\setcounter{table}{0}
\setcounter{figure}{0}

\maketitlesupplementary


\section{Video Detection \& Sparsification Results}
In the accompanying multimedia file, titled \textbf{\texttt{video.mp4}}, we present visualizations corresponding to several event clips in the test set of 1Mpx. 
This video includes comparisons between ground truth and SAST’s object detection results, showcasing SAST's high performance.
It also features the visualizations of score heatmaps and selection results across different scenes, providing a clear demonstration of SAST's scene-aware adaptability. From the video, it can be observed that SAST assigns higher scores to important tokens within important windows and performs a series of operations such as self-attention, MLP, and normalization exclusively on these sparse tokens, significantly reducing computational costs.

\section{Additional Experiments}
\subsection{Sparsity Level of SAST.}
\begin{figure}[h]
\centering
\includegraphics[width=\columnwidth]{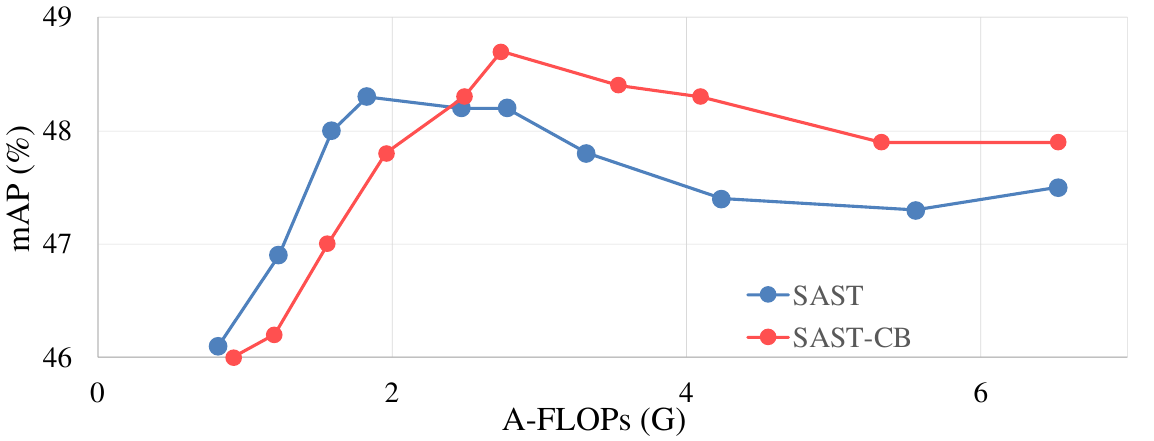}
\caption{ Setting different hyper-parameters results in different sparsity levels and performance of SAST and SAST-CB on 1Mpx. Performance does not continuously improve with increasing sparsity levels. SAST and SAST-CB each have their advantages in sparser and denser settings.}
 \vspace{-3mm}
\label{fig_ablsparse}
\end{figure}

We adjust the hyper-parameters $a$ and $b$ to limit the sparsification of SAST and SAST-CB, resulting in 10 sparsity levels. The performance of 20 networks is illustrated in \cref{fig_ablsparse}. 
SAST and SAST-CB respectively excel at sparser and denser sparsity levels, but both achieve the best results at a moderate sparsity level. We interpret these results from an information perspective, where higher information density (more effective interactions among fewer tokens) has been shown to benefit the Transformers \cite{hyeon2022scratching}. 
SAST-CB achieves this by broadcasting information among selected tokens, enhancing effective information interactions within a reduced token set.
However, an overly sparse network can lead to information loss due to excessive compression. Therefore, the choice between SAST and SAST-CB depends on the specific requirements of the task at hand: SAST for lower computational load and SAST-CB for enhanced detection at a slightly higher computational cost.

\begin{table}[h]
  \centering
  \tabcolsep=0.03cm
  \begin{adjustbox}{width=\columnwidth,center}
    \begin{tabular*}{\hsize}{@{}@{\extracolsep{\fill}}l|cc|cc@{}}
      \multicolumn{1}{c}{} 
      & \multicolumn{2}{c}{1Mpx} 
      & \multicolumn{2}{c}{Gen1} \\
      \cline{1-5}
      Functions
      & mAP (\%) & A-FLOPs (G) & mAP (\%) & A-FLOPs (G)
      \\
      \hline
      Identity
      & 46.5 & 2.3 & 46.1 & 1.2\\
      SoftMax 
      & 40.9 & 1.5 & 42.0 & 0.7\\   
      Tanh
      & 47.0 & 1.7 & 46.7 & 0.8\\
      Sigmoid
      & \textbf{48.3} & 1.8 & \textbf{47.9} & 0.8 \\
      \hline
    \end{tabular*}
  \end{adjustbox}
  \vspace{1mm}
  \caption{Detection performance on 1Mpx and Gen1 using different weighting functions. Sigmoid achieves the optimal results.}
  \vspace{-3mm}
  \label{tab_abl3}%
\end{table}

\begin{table*}[t]
  \centering
  \tabcolsep=0.098cm
  \begin{adjustbox}{width=2\columnwidth,center}
    \begin{tabular*}{\hsize}{@{}@{\extracolsep{\fill}}lc|cc|cc|c@{}}
      \multicolumn{2}{c}{} 
      & \multicolumn{2}{c}{1Mpx} 
      & \multicolumn{2}{c}{Gen1} 
      &  \\
      \cline{1-7}
      Methods
      & Backbone
      & mAP (\%) & FLOPs (G) 
      & mAP (\%) & FLOPs (G) 
      & Params (M)
      \\
      \hline

      RVT-L \cite{rvt} & MaxViT-L \cite{maxvit}
      & 47.8 & 23.2\ (19.3) 
      & 47.6 & 7.8\ (6.6) 
      & 33.2 \\
      Ours & SAST-L
      & \textbf{49.2} (\textcolor{ForestGreen}{+1.4})& \ \ \textbf{7.5}\ (\textbf{3.7},\ \textcolor{ForestGreen}{-81$\%$}) 
      & \textbf{48.6} (\textcolor{ForestGreen}{+1.0})& \textbf{2.9}\ (\textbf{1.7},\ \textcolor{ForestGreen}{-74$\%$}) 
      & 33.6 \\

      \hline
    \end{tabular*}
  \end{adjustbox}
 \vspace{-2mm}
  \caption{
  Detection performance on 1Mpx and Gen1 by training larger variants of RVT and SAST.}
 \vspace{-2mm}
  \label{tab_deep}%
\end{table*}





      

\subsection{Weighting Method Ablation.}
In \cref{tab_abl3}, we conduct a comparative analysis of various functions used in the STP weighting process for transitive derivatives. The Sigmoid function, which smoothly maps scores to weights within the range of 0 to 1, assigning higher weights to more significant tokens, delivers superior performance on both datasets.

\subsection{Bigger Model, Bigger Gain.}
We scale up the RVT and SAST by increasing the layer count in the first, second, and fourth blocks by a factor of two and in the third block by a factor of six, producing the larger RVT-L and SAST-L variants. As depicted in \cref{tab_deep}, training these larger models on the 1Mpx and Gen1 datasets both result in performance gains. 
However, for RVT-L, the proportion of A-FLOPs within the total FLOPs significantly increases, which echoes the discussions of model scalability in \cref{sec:intro}. At a comparable model size, SAST-L exhibits more pronounced benefits from its adaptive sparsification. It achieves an impressive mAP of $49.2\%$ on 1Mpx and $48.6\%$ on Gen1 datasets, with even fewer FLOPs than the pre-scaled RVT-B. This fully demonstrates the potential of our proposed sparsification method in achieving a remarkable balance between performance and computational cost for large models.

\section{Additional Visualizations}
\begin{figure*}[b]
\centering
\includegraphics[width=1.75\columnwidth]{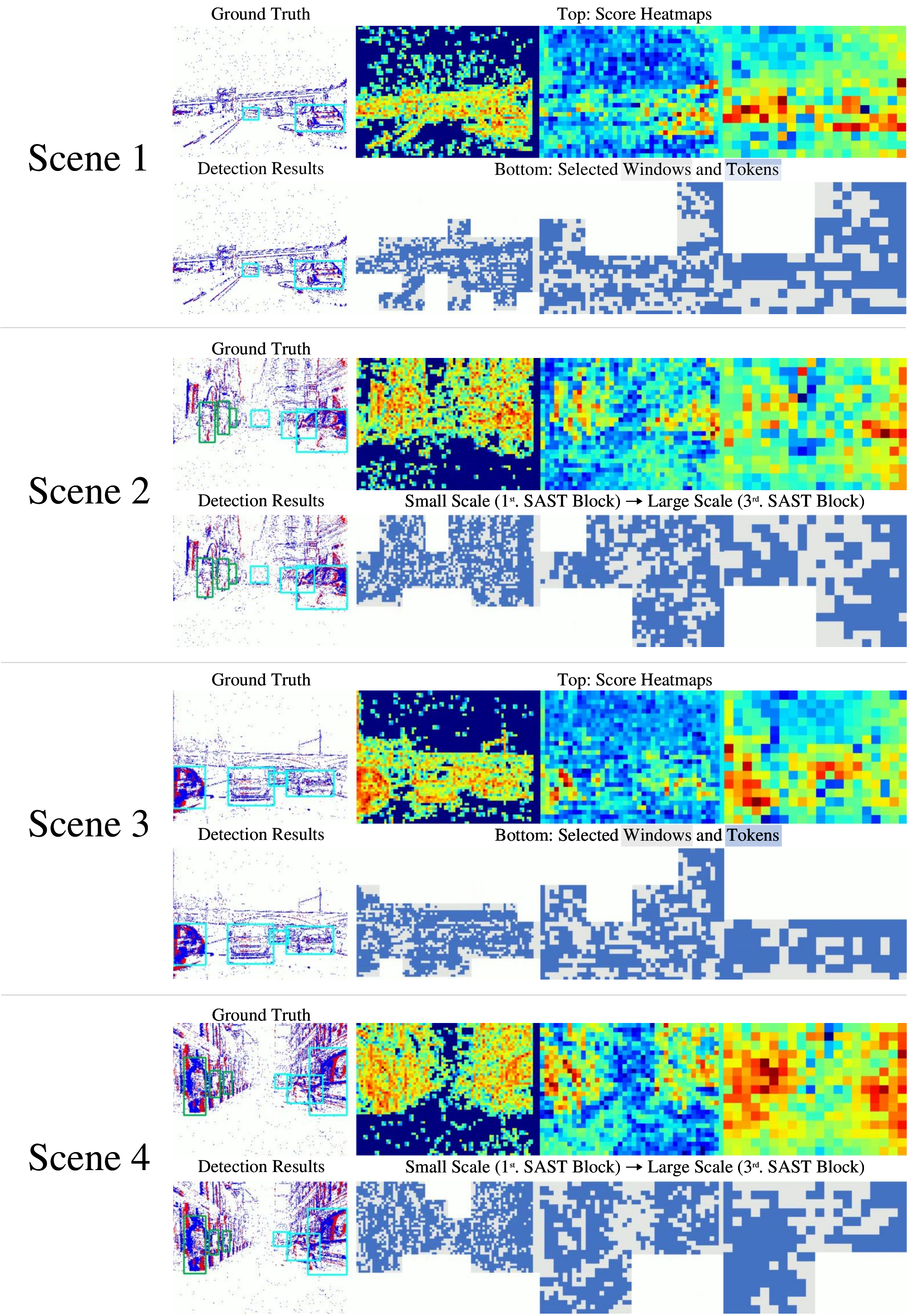}
 \vspace{-1mm}
\caption{Additional Visualizations of original events, score heatmaps, and selection results under four scenes in Gen1. 
As the network progresses through subsequent SAST blocks, featuring multiple downsampling stages, the scale (receptive field) of tokens expands.
}
 \vspace{-2mm}
\label{fig_visualizations_gen1}
\end{figure*}

\begin{figure*}[b]
\centering
\includegraphics[width=1.75\columnwidth]{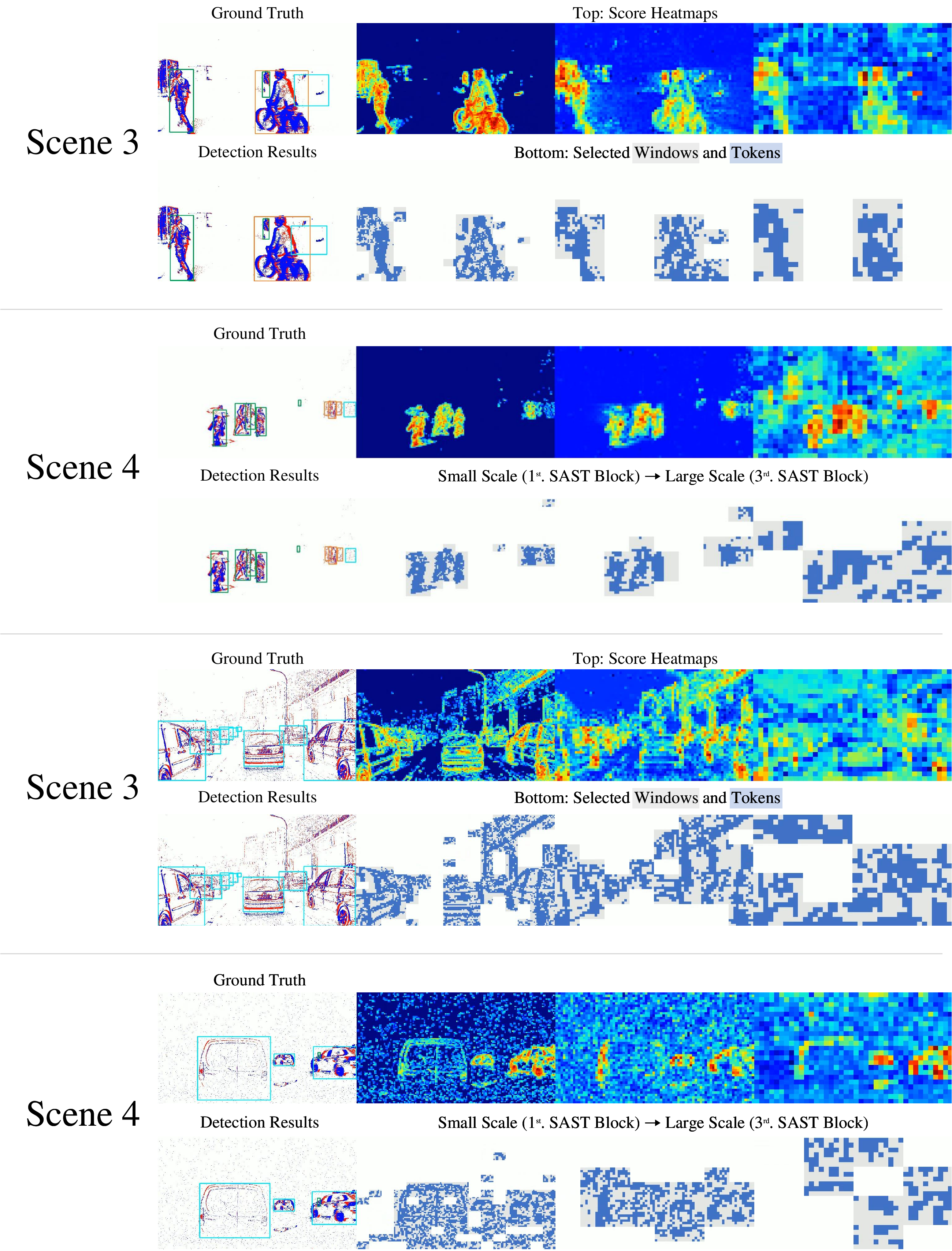}
 \vspace{-1mm}
\caption{Additional Visualizations of original events, score heatmaps, and selection results under four scenes in 1Mpx. 
As the network progresses through subsequent SAST blocks, featuring multiple downsampling stages, the scale (receptive field) of tokens expands.
}
 \vspace{-2mm}
\label{fig_visualizations_1mpx}
\end{figure*}

We extend our visualization analysis to the 1Mpx and Gen1 datasets, as shown in \cref{fig_visualizations_gen1} and \cref{fig_visualizations_1mpx}. These visualizations encompass the original event data, score heatmaps, and the selection results of the windows and tokens. The supplementary visualizations reinforce our findings from the main text, providing further evidence of the network's scene-aware adaptability in assigning higher scores to important tokens and adjusting sparsity levels in response to the scene complexity.

\section{Additional Implementation Details.}
In \cref{tab_para}, we list the default choices of key hyper-parameters, facilitating the replication of our study. These parameters are shared for both the 1Mpx and Gen1 datasets.
\begin{table}[h]
  \centering
  \tabcolsep=0.03cm
  \begin{adjustbox}{width=\columnwidth,center}
    \begin{tabular*}{\hsize}{@{}@{\extracolsep{\fill}}cccccc@{}}

      \cline{1-6}
      
      $a$ & $b$ & $p$ & Batch Size & Steps & Learning Rate
      \\
      \hline
      0.0002 & 0.099 & 1.0 & 32 & 600000 & 0.00056\\
      \hline
    \end{tabular*}
  \end{adjustbox}
  \vspace{1mm}
  \caption{Default hyper-parameters used for training SAST and SAST-CB on 1Mpx and Gen1.}
  \vspace{-3mm}
  \label{tab_para}%
\end{table}

\end{document}